\documentclass[letterpaper, 10 pt, conference]{ieeeconf}  

\IEEEoverridecommandlockouts                              

\usepackage{cite}
\usepackage{amsmath,amssymb,amsfonts}
\usepackage{algorithmic}
\usepackage{graphicx}
\usepackage{textcomp}
\usepackage{subcaption}
\usepackage{romannum}
\usepackage{url}

\usepackage{float}
\usepackage{multirow}

\title{\LARGE \bf
The Power of Color: A Study on the Effective Use of Colored Light in Human-Robot Interaction
}

\author{Aljoscha P\"{o}rtner$^{1}$$^{,}$$^{2}$, Lilian Schr\"{o}der$^{1}$, Robin Rasch$^{1}$$^{,}$$^{3}$, \\ Dennis Sprute$^{1}$$^{,}$$^{2}$, Martin Hoffmann$^{1}$ and Matthias K\"{o}nig$^{1}$
\thanks{*This work is financially supported by the German Federal Ministry of Education and Research (BMBF, Funding number: 03FH006PX5).}
\thanks{$^{1}$Bielefeld University of Applied Sciences, Campus Minden, 32427 Minden, Germany, Email: {\tt\small surname.name@fh-bielefeld.de}}%
\thanks{$^{2}$Otto-von-Guericke University Magdeburg, Faculty of Computer Science, 39106 Magdeburg, Germany}%
\thanks{$^{3}$Bielefeld University, 33619 Bielefeld, Germany}%
}

\begin{document}

\maketitle
\thispagestyle{empty}
\pagestyle{empty}

\begin{abstract}

  In times of more and more complex interaction techniques, we point out the powerfulness of colored light as a simple and cheap
  feedback mechanism. Since it is visible over a distance and does not interfere with other modalities, it is especially interesting for mobile robots.
  In an online survey, we asked 56 participants to choose the most appropriate colors for scenarios that were presented in the form of videos. In these scenarios a mobile robot accomplished tasks, in some with success, in others it failed because the task is not feasible, in others it stopped because it waited for help. We analyze in what way the color preferences differ between these three categories. The results show a connection between colors and meanings and that it depends on the participants' technical affinity, experience with robots and gender how clear the color preference is for a certain category. Finally, we found out that the participants' favorite color is not related to color preferences.

\end{abstract}

\section{Introduction}
By using intuitive techniques and multiple modalities, modern user interfaces become more and more flexible. While natural interaction techniques like gestures or speech recognition are still subject to development, this paper considers light as a simple and rather traditional, but yet reliable feedback mechanism. Throughout household and industrial applications, light in many variations is used to indicate the inner state of devices. The expressive power of light is founded on several dimensions of interaction: duration, sequential combinations of durations, brightness and color. In order to systematically examine the different dimensions, the objective of the following study is to explore how the light color affects and increases the intelligibility of status indication in robotics applications. One important question is whether the color of a light signal can convey a meaning by association, not only by being learned. Related work regarding colored light in human-robot interaction is very limited and more general studies show varying results. For this reason, we systematically evaluated whether there are color-meaning associations by conducting an online video survey with 56 participants\footnote[4]{The videos can be found at: \url{https://www.youtube.com/watch?v=AsWSeUI3hbI&list=PLa0JeK8XpdpCM4gCl03Miu4qvBcDGQnyU}}.
The following section considers related work concerning lights in human-computer and human-robot interaction and points out why they are of interest for the design of mobile robots. Afterwards, the design of the study as well as its results and implications and future research directions are presented.

\begin{figure*}
\begin{subfigure}{0.245\textwidth}
\includegraphics[width=1\textwidth]{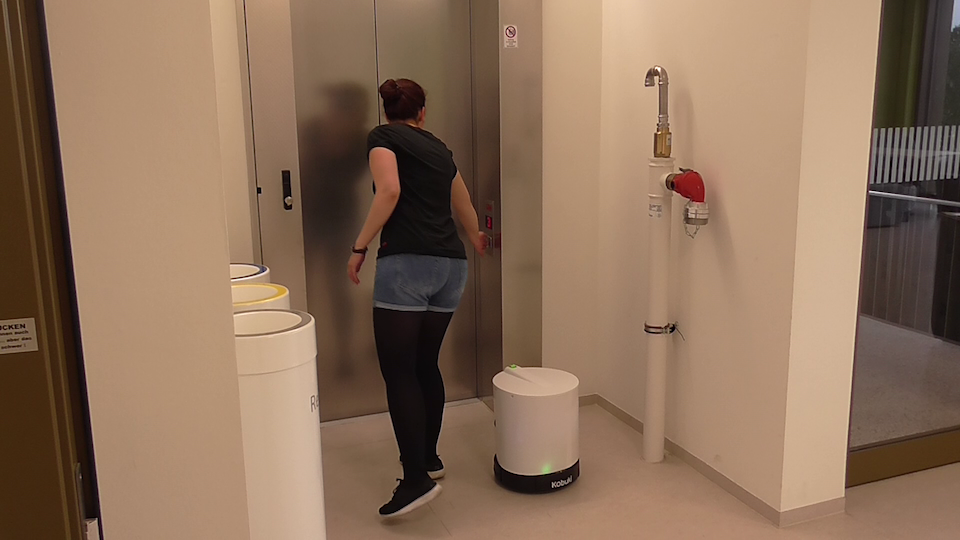}
\label{fig:figure1a}
\subcaption{Elevator}
\end{subfigure}
\begin{subfigure}{0.245\textwidth}
\includegraphics[width=1\textwidth]{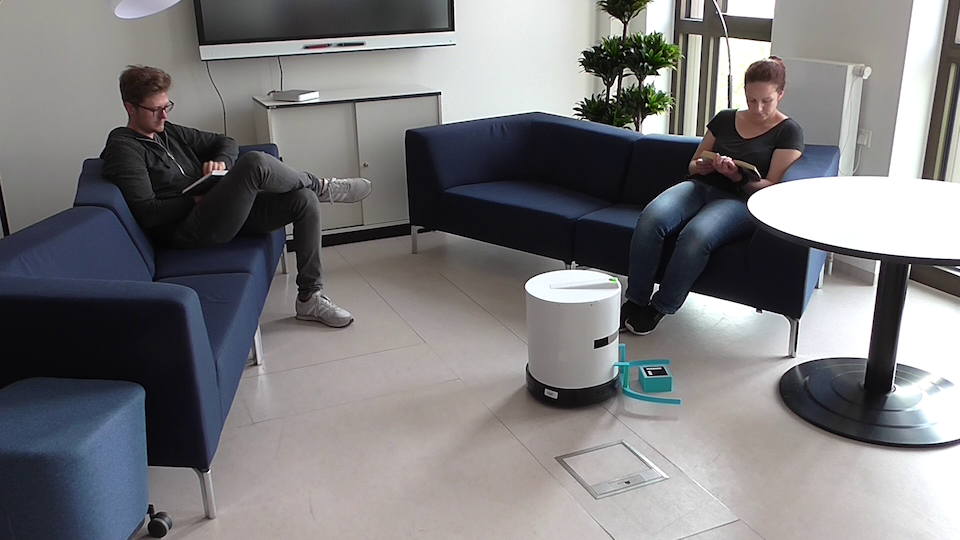}
\label{fig:figure1b}
\subcaption{Seating area}
\end{subfigure}
\begin{subfigure}{0.245\textwidth}
\includegraphics[width=1\textwidth]{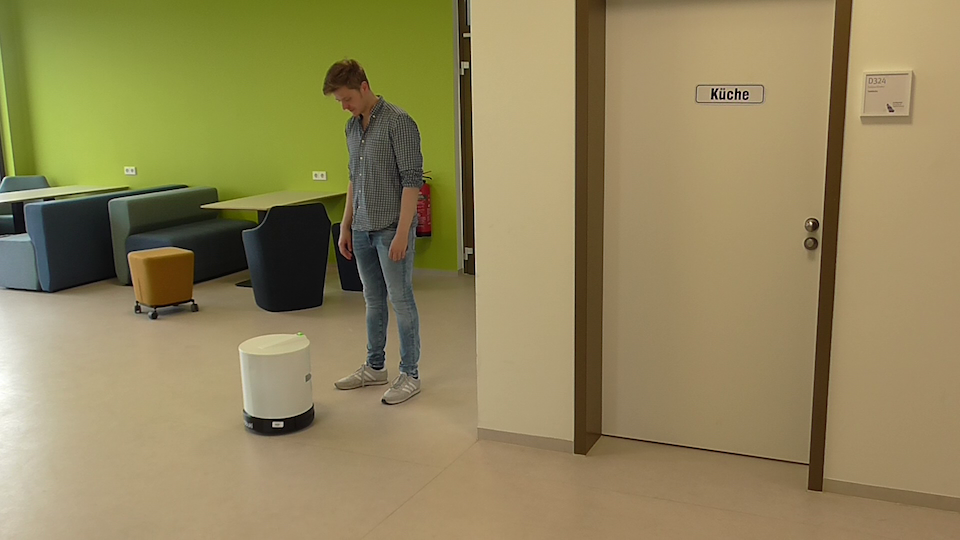}
\label{fig:figure1c}
\subcaption{Hallway}
\end{subfigure}
\begin{subfigure}{0.245\textwidth}
\includegraphics[width=1\textwidth]{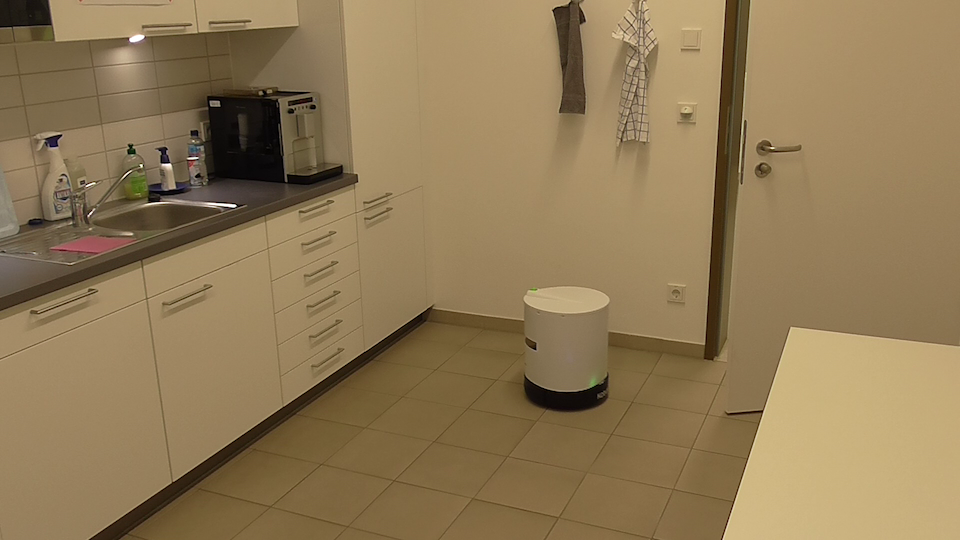}
\label{fig:figure1d}
\subcaption{Kitchen}
\end{subfigure}
\vspace{-0.8em}\caption{Illustration of different scenarios addressed by the study}\label{fig:figure1}
\end{figure*}
\section{Related Work}
The specific question in what way colored light is suitable for effective feedback of mobile robots is quite new, therefore we relied on related work from more general fields. Some previous studies examine the general psychological effect of color on people while others investigate also the practical utilization of light and color as methods of interaction between humans and computers or humans and robots.
The following subsections discuss these topics in detail.

\subsection{The Meaning of Colors}
Many studies have tried to explore the psychological effect of color on people. Various studies examine the relationship between colors and emotions. Terwogt and Hoeksma~\cite{terwogt1995colors} examine the emotion-color linkages of three age groups from child to adult. Their evaluation shows that the linkages shift over time, e.g. seven year old children tend to relate blue with happiness, while adults tend to relate green with happiness. Their study also reveals that at all ages colors and emotions are strongly related. The same conclusion is drawn by Dael et al.~\cite{dael2015} who investigate the linkage between color and the bodily expressions of elated joy and panic fear. Within their investigation, the users tend to choose colors of the red to yellow range of the hue encoding in case of elated joy and colors of the cyan to blue range in case of fear. In general, different studies with varying results imply that the association between colors and emotions is very heterogeneous. While Palmer et al.~\cite{palmer2013} reveal that red color tones are associated with both happiness and anger, other studies yield associations between blue and positive feelings~\cite{valdez1994,francis1973} as well as calm and sad emotions~\cite{palmer2013}. Naz and Helena~\cite{naz2004relationship} state that principle hues (red, yellow, blue etc.) are more connected to positive emotional responses than intermediate hues. Apart from general psychological aspects and interpersonal differences, there might be a contextual influence on color perception. For example, the interpretation of a red light on a TV is probably different from the interpretation of a red lamp on a car's instrument panel or a synthetic light in a laboratory environment. Therefore, the following section reviews the related work regarding color and light in human-computer as well as human-robot interaction in order to account for this contextual influence.

\subsection{Colored Lights in Human-Computer and Human-Robot Interaction}
Most devices of everyday use contain one or more lights that indicate information about the device. This includes devices all over the household like kitchen devices as well as multimedia devices or smartphones. Even though the use of lights is widespread, by far not all possibilities of this modality have been used, evaluated and studied yet. In order to simulate emotions, the field of affective computing examines the possibility to incorporate color and light in the interaction process to anthropomorphize computers. A review on the role of color and light in affective computing is given by Sokolova and Fernandes-Caballero~\cite{sokolova2015}. Pradana et al.~\cite{pradana2014} study the possibility to prime emotions while reading incoming text messages using a ring-shaped color lighting wearable.
Based on the assumption that there are positive and negative colors, the study shows that the use of such colors can shift the user's emotional reaction on incoming messages. Song and Yamada~\cite{song2017a} conducted a preliminary study that shows a positive effect of LED light animations on users during a computer game in terms of anthropomorphism and the user's perception of the computer.
While the use of color and light in affective computing is used to simulate emotions, the use of color and light to illustrate the status of a device is more common on a daily basis. An overview on how lights are used as a feedback modality in various technical devices is given by Harrison et al.~\cite{harrison2012unlocking}. Their study suggests that there are connections between light behaviors, e.g. blinking animations of LEDs, and the conveyed meaning. The systematic and empirically meaningful use of lights and colors to illustrate device states in robotics is rare. Sprute et al.~\cite{sprute:2017} use colored light as a feedback mechanism in robotics. The current state of the system is illustrated using an LED strip mounted on the robot and using colored smart lights of the environment. Baraka et al.~\cite{baraka2016expressive} recommend different color and light animation compositions for different robot scenarios, e.g. \textit{waiting}, \textit{blocked} and \textit{progress}. In order to identify the best animation for each scenario, they carried out a study where experts and naive users were asked to give input on the animation type, speed and color. The result of this work was a list with a detailed description of the best animations for the introduced scenarios. Based on the previously mentioned study, Baraka et al.~\cite{baraka2016enhancing,baraka2017mobile} later investigate the positive effects of colored light on the understanding of and trust in robots by humans. They conducted a study where users hypothesize about the robot's internal state in different robot scenarios supported by colored lights. In order to compare the different scenarios and to make a more universal statement, they categorize them in three distinct scenario classes. The study shows that the use of light has a significant effect on the accuracy of the users hypothesis as well as the general trust in robots. They claim that light signals are intuitively understandable, because humans generalize from other experience with devices that use light signals. Szafir et al.~\cite{szafir2015} propose different visual signaling mechanisms for flying robots using LED strips. They evaluate whether different light animations can help to enable users to deduce flight intents. The evaluation reveals that, while the participants believed that all animations help to make the robot a better partner, only a subset of animations enable participants to understand the robot's light intents more quickly and accurate. A similar approach is proposed by Monajjemi et al.~\cite{monajjemi2016}. Betella et al.~\cite{betella2013} conduct a study to find how affective states of technical devices can be expressed using colored motifs. The study reveals that the perception of affective states of technical devices is correlated to the color of the motif, e.g. pleasure is correlated to cool colors (long-wavelength hues).  Collins et al.~\cite{collins2015} illustrate the internal state of a robot using changing patterns of light based on the circumflex model~\cite{posner2005}. The results of their study show that changing patterns of light can evoke reliable perceptions of affect in naive participants and can therefore convey complex information.

\section{Scenario and Approach}
In our scenario, a mobile robot was able to interact and communicate with users using an attached LED strip shining in different colors. The mobile robot was located in a building and performed various scenarios. It was able to react to complex speech commands, to navigate on its own, to clean a room and was able to perceive objects in its environment. The study was designed as a \textit{Wizard of Oz experiment} where the behavior was realized by simply remote controlling the robot. Real abilities were not necessary in our case, since participants only saw recorded videos of the robot. 
\begin{figure*}
\centering
\includegraphics[width=1\textwidth]{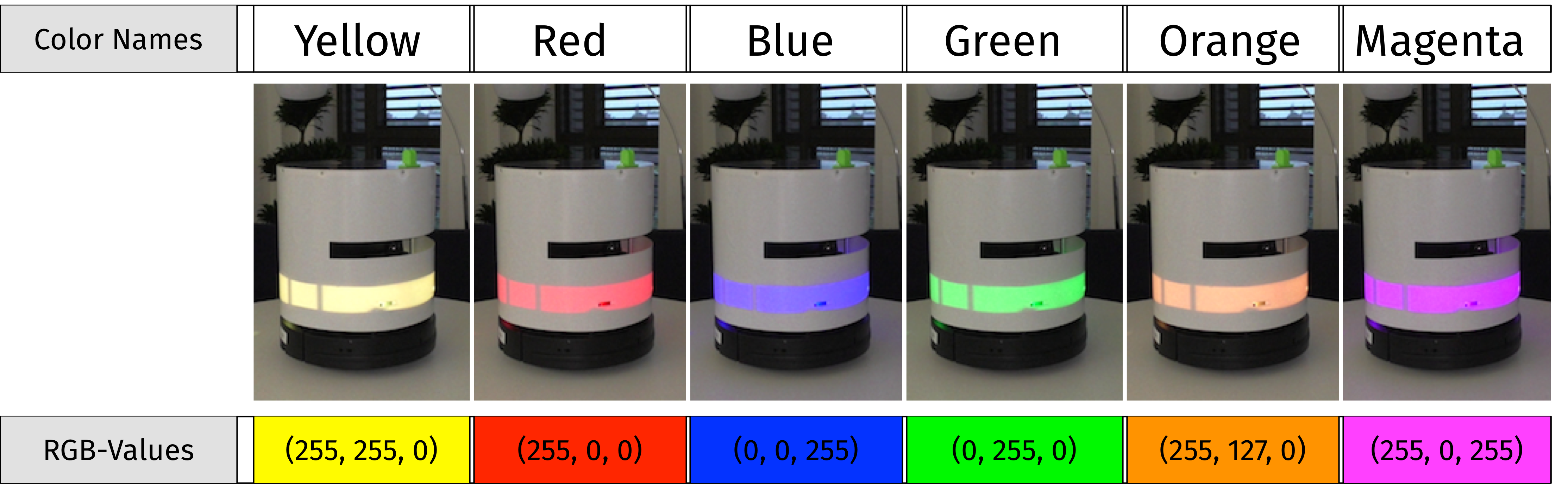}
\caption{Mobile robot with different light coloring}\label{fig:2}
\end{figure*}
The videos were recorded at four different locations in a university building: in front of an elevator~(Fig.~\ref{fig:figure1}a), at a seating area in a laboratory~(Fig.~\ref{fig:figure1}b), in a hallway~(Fig.~\ref{fig:figure1}c) and in a kitchen~(Fig.~\ref{fig:figure1}d).
The LED strip is a fully-addressable \textit{WS2812-based} RGB LED strip with \textit{30} LEDs per meter. The strip is \textit{12.5 mm} wide, \textit{4 mm} thick and is connected to an Arduino UNO using a \textit{2-pin JST SM connector}. It is mounted on a TurtleBot2 with a white custom casing made of thin polystyrene in order to enable translucency~(Fig.~\ref{fig:2}).
\section{User Study}
We conducted an experimental evaluation using an online survey to reveal the relationships between colored light and its semantics in the context of robotics. We assumed that the experience with colored light in daily life might influence our color semantics. This would also mean that the association is shaped by the degree and nature of the experience, which varies from person to person. This could lead to varying color semantics. Therefore, it was interesting to determine how experienced a user is.
Our main question was whether there are stable prototypes of colors' semantics that play a role in human-robot interaction or not. The study was conducted with a total of 56 participants, out of which 38 were male and 18 were female. Their ages ranged from 15 to 70 years (Mdn = 26, SD = 13.7), only one of them belonged to the group of elderly people (\textgreater 60 years).
\subsection{Experimental Design}
The study was designed to reveal the relationship between colored light and its semantics in robotics. The participants saw videos of a mobile robot in different situations and were supposed to select the color that would help them to assess the situation and understand the state of the robot. In this context, it was assumed that every state of a robot is connected with an information the robot wants to transmit to the user. Following the work by Harrison et al.~\cite{harrison2012unlocking} and Baraka et al.~\cite{baraka2016expressive}, we identified three categories of states (information) that are general enough to cover several specific states. The three categories are \textit{Active}, \textit{Help needed} and \textit{Error}, and they are shown in Table~\ref{tab:1}. According to Baraka et al.~\cite{baraka2016expressive}, we specified scenarios for each category to make them more tangible for our participants than abstract categories.
\begin{table}
\centering
\begin{tabular}{|p{1cm} | p{2cm} | p{4.3cm}|} \hline
  \textbf{Index} & \textbf{Information \mbox{category}} & \textbf{Description}  \\ \hline
\textit{\Romannum{1}.}) & Active & The system is computing or progressing.  \\
\textit{\Romannum{2}.}) & Help needed & The system is unable to proceed without help / the system is waiting. \\
\textit{\Romannum{3}.}) & Error & The system is unable to connect or to accept input or to proceed with its task. \\
\hline
\end{tabular}
\caption{Detailed description of scenario classes}
\label{tab:1}
\end{table}
In total, the participants saw 9 videos, each representing a specific scenario, e.g. a robot searching for an object. All scenarios can be categorized in three distinct scenario classes: \textit{Delivery tasks}, \textit{Navigation tasks} and \textit{Cleaning tasks}. The example scenarios, their classification as well as their relationship to the information categories are shown in Table~\ref{tab:2}.
We decided to represent the scenarios with the help of videos rather than a textual description. This is closer to the real-world situation that a human observes a robot acting. In the videos, the robot received speech commands because this way the task was clear and unambiguous. The speech was also represented textually as a subtitle, in case the participant had a hearing loss.
\newline
\begin{table}
\centering
\begin{tabular}{ | p{1.3cm} |l| p{5.5cm} |} \hline
\centering
  \textbf{Information category} & \textbf{Index} & \textbf{Example scenario}\\ \hline
 \multirow{3}{2cm}{Active} & \textit{\Romannum{1}.a}) & Delivery task: a user tells the robot to deliver an object, the robot searches and finds it and starts to drive back to the user with the object.\\ \cline{2-2}
 & \textit{\Romannum{1}.b}) & Navigation task: a user tells the robot to go to a specific room. The robot navigates through the building, the video stops shortly before the goal is reached. \\ \cline{2-2}
 & \textit{\Romannum{1}.c}) & Cleaning task: The robot is told to clean the kitchen. It starts doing this, the video stops before it is finished. \\
 \hline
\multirow{3}{2cm}{Help needed} & \textit{\Romannum{2}.a}) & Delivery task: Same situation as in \textit{Active} category, but the robot needs help because the object is out of reach. The video stops when the robot is in front of a table. A short sequence before the actual video shows a similar situation where a human helps the robot by putting the object on the floor. \\ \cline{2-2}
 & \textit{\Romannum{2}.b}) & Navigation task: Same situation as before, but the robot cannot pass a closed door. The video stops when it is in front of the door. A short sequence before the actual video shows a similar situation where a human opens the door.\\ \cline{2-2}
 & \textit{\Romannum{2}.c}) & Navigation task: same situation as before, but robot cannot open the elevator.
  The video stops when it is in front of the elevator. A short sequence before the actual video shows a similar situation where a human presses the elevator button.\\
\hline
\multirow{3}{2cm}{Error} & \textit{\Romannum{3}.a}) & Delivery task:  Same situation as in \textit{Active} category, but the object is not available. The video shows the empty room and that the robot searches in it. The robot stops searching at the end of the video.\\ \cline{2-2}
& \textit{\Romannum{3}.b}) & Navigation task: Same situation as before, but robot is stuck in front of a ``No access'' sign on the kitchen door. \\ \cline{2-2}
& \textit{\Romannum{3}.c}) & Cleaning task: The robot is told to clean ``behind the couch''. It drives towards the couch from two directions, but it is clear to see that there is no way around the couch and the robot stops in front of it.\\
\hline
\end{tabular}
\caption{Detailed description of scenarios}
\label{tab:2}
\end{table} We ensured that the categories \textit{Help needed} and \textit{Error} could be distinguished: If there was a possible solution to the robot's problem, the video included a short sequence showing this solution. For example, in one video the robot was supposed to navigate to another floor of the building. At the beginning of the video, a person opened the elevator for the robot which clarified that this is a situation where a human can act as a supporter. The video then was rewound to the point before the helping person was visible.

Due to practical reasons, 
the wide spectrum of colors was pooled together. We assume that this can be done because a color-meaning association includes rather color prototypes than subtle color variations.
Previous studies already addressed the question how this pooling should be conducted. In the study by Baraka et al.~\cite{baraka2016expressive}, the participants chose the most suitable light for one scenario out of a range of six distinct colors. The colors they used were orange, green, soft blue, dark blue and purple.
We preferred color options that cover the whole color spectrum. There are many different ways to categorize colors in color theory. We selected colors with a high contrast, both the primary colors (yellow, red, blue) and the secondary colors. The resulting colors were, with RGB-values in brackets: \textit{red} (255,~0,~0), \textit{blue} (0,~0,~255), \textit{yellow} (255,~255,~0), \textit{green} (0,~255,~0), \textit{magenta} (255,~0,~255) and \textit{orange} (255,~127,~0). The representation of the colors using the mobile robot is illustrated in Fig.~\ref{fig:2}.
%
\subsection{Method}
We conducted an online study in order to achieve a higher number of participants. At the beginning of the survey, participants read detailed instructions. Using Google Forms, we were able to conduct an online survey with embedded videos and all necessary answer and question formats. Participants only had to follow a certain link, the survey itself included all necessary information and instruction, which is beneficial for standardization.
After the introducing instruction, a first video showed the robot with light on, colored in the six available colors, one after the other. The colors were accompanied with a verbal color description (``red'', ``green'' etc). Even though the persons were not directly interacting with the robot yet, they gained an intuitive impression of the light signal directly on the target light hardware.
The robot accomplished the previously mentioned service robot tasks, and its appearance was dominated by a neutral white housing that allowed the colors to be bright and clear. We assumed that a robot with a simple, round form is a good platform for our question. On a humanoid robot, for example, the position of the lights might influence their perception as the light signal could be related to a body part or one specific function. The TurtleBot's compact form probably appears less ambiguous.
After introducing the robot in different colors, they viewed short videos that showed certain situations where the robot was more or less successful. After each video the participants were asked to select a suitable color to represent the current internal state of the robot. The scenarios were designed to contain one relevant information category, e.g. ``the robot needs help''.  The introducing instructions included the information about the three categories.
For every information category the participants saw three different videos, so every participant saw nine videos.
At the end, the survey contained questions concerning the level of computer experience and experience with robots as well as demographics. These questions were placed after the color choices so they did not affect them.
\subsection{Hypotheses}
The study investigates the following hypotheses:
\begin{itemize}
  \item[\textbf{H1}] There are stable prototypes of color semantics.
  \item[\textbf{H2}] Participants choose their favorite color more often than other colors.
  \item[\textbf{H3}] The individual preference of colors is influenced by the technical affinity.
  \item[\textbf{H4}] The individual preference of colors is influenced by the experience with robots.
  \item[\textbf{H5}] The individual preference of colors is not influenced  by gender.
\end{itemize}
If a color is associated with a semantic meaning in a stable way, colors are not equally suitable in a certain scenario. If consistent differences in the color preferences occur, this is evidence for \textbf{H1}.
This connection between color and meaning would be distorted by the favorite color, if \textbf{H2} is supported by our results. Moreover, a person's experience with robots or technical affinity might influence the color choice (\textbf{H3} and \textbf{H4}). For example, a lack of experience could lead to a less systematic color choice because the connection between color and meaning could not be developed. If there is a connection, it is probably the result of repeated experiences of joint appearance of a color and a meaning. Finally, we did not expect the person's gender to influence the color choice~(\textbf{H5}).
\subsection{Metrics}
We measured the color preference, i.e. which colors each participant chose for the three categories. Since every person had to choose three times for each category, we were also interested in how consistently they chose the colors. To test our hypotheses, we measured additional variables: the technical affinity, the extend of robot experience, age and gender. As we explained before, the experience with technical systems probably influences the perception of signals and symbols. The questions about technical affinity were taken from a sophisticated German questionnaire developed to control for this variable in human-machine interaction research~(TA-EG)~\cite{karrer2009technikaffinitat}. In order to ensure that our results are significant, we performed $\chi^2$ goodness-of-fit tests against uniform distributions. Furthermore, to test if the results are significantly different among two groups, we performed $\chi^2$ homogeneity tests.
In both cases, the significance level is defined to be $\alpha = 0.05$.
\section{Results}
\begin{figure}[H]
\centering
\includegraphics[width=0.4\textwidth]{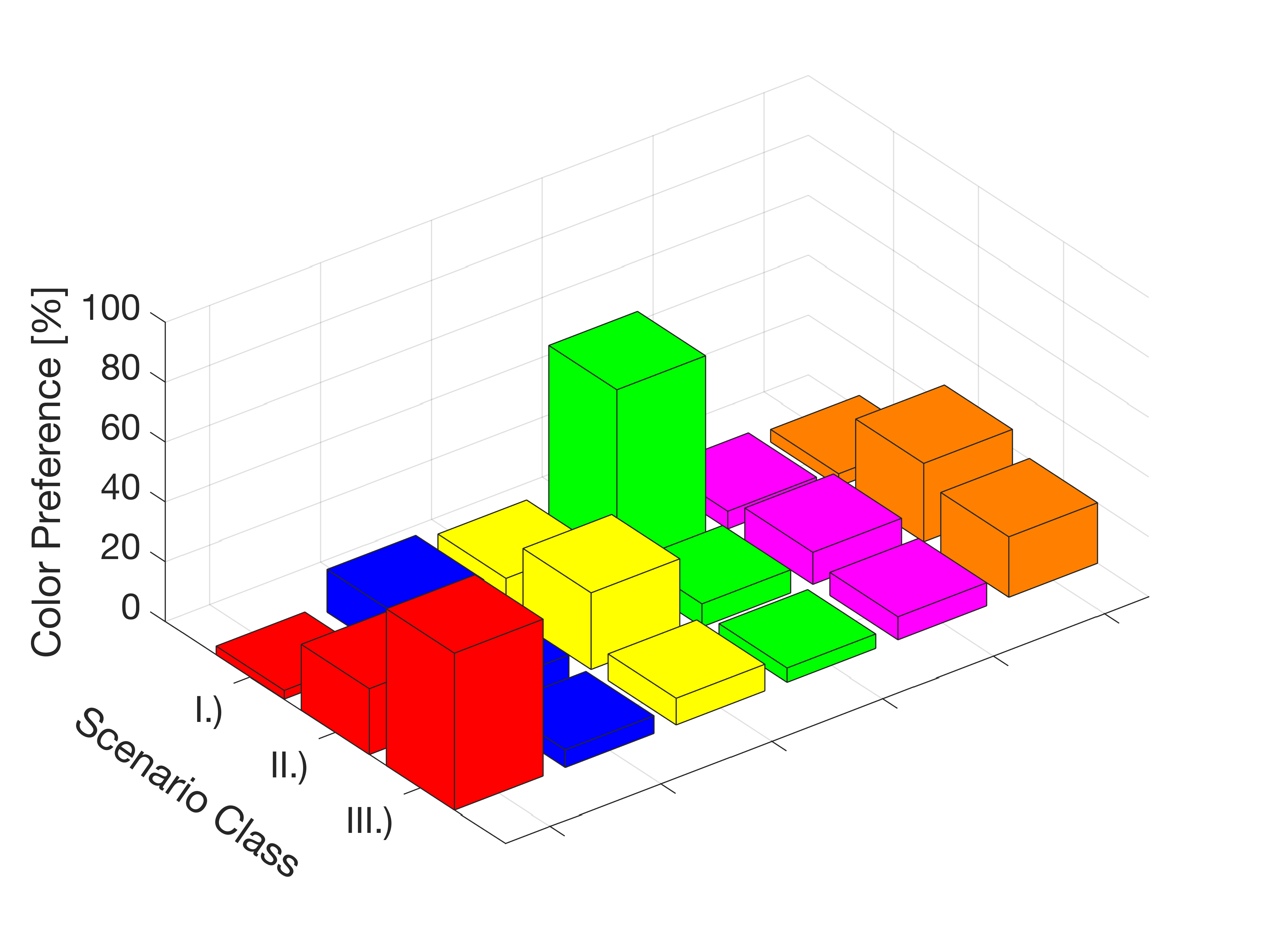}
\caption{All participants}\label{allcol}
\end{figure}

Fig. \ref{allcol} shows the color preferences of all participants. The scenario classes are labelled according to the indices in Table~\ref{tab:1}. At first, we performed a $\chi^2$ goodness-of-fit test against a uniform distribution. For all three categories, the results are highly significant.
The preference is not equally distributed in the groups, for the \textit{Active} category $\chi^2(5, N=168)=244.64$, for the \textit{Help needed} category $\chi^2(5, N=168)=39.71$, for the \textit{Error} category $\chi^2(5, N=168)=169.79$, $p\textless.01$ in all cases. This shows clearly that the color choice is not random and that some colors are considered more suitable for a category than others. Nevertheless, this only leads to a clear ``recommendation'' for \textit{one} color in two out of three cases. Secondly, we were interested in how consistent the parti\-cipants voted. The more consistent their choice, the more we can be sure that they associate a color with a meaning.
Out of all participants, 89.3\% voted the same color two or three times in the \textit{Active} category,  83.9\% in the \textit{Help needed} category and 78.6\% in the \textit{Error} category. Furthermore, we analyzed how many participants chose the same color every time: 42.9\% in the \textit{Active} category, 46.4\% in the \textit{Help needed} category and 39.3\% in the \textit{Error} category. So the great majority of participants chose a color for a category at least twice. For all participants, there seems to be a clear preference for the categories \textit{Active} (green with over 60\%) and \textit{Error} (red with more than 50\%). However, for the category \textit{Help needed} there are three colors with similar percentages: red, orange and yellow.
For a more profound analysis, we examined the influence of four possible confounding variables: the participants' favorite color, their technical affinity, their experience with robots and their genders.
In order to examine the effect of the participants' favorite color on their selection, we counted how many persons chose their favorite color two or three times, depending on the category:
26.8\% (\textit{Active}), 1.8\% (\textit{Help needed}) and 14.3\% (\textit{Error}). The most frequent favorite color was blue, which had low ratings for all categories.

To control for the variable technical affinity, we divided the participants in two groups according to their rates on the affinity dimensions and performed individual tests. The questionnaire provided rates on four dimensions (competence, enthusiasm, negative attitude and positive attitude). In order to reduce this to an dichotomous value, our operationalization was as follows: People were considered to have a high affinity if they had high ratings in more than two dimensions\footnote{High rating means less than three on a five-point Likert scale where one is the highest rating. The rating for negative attitude was inverted before.}. 22 participants had a low technical affinity, 34 a high technical affinity. 
\begin{figure}
\centering
\includegraphics[width=0.5\textwidth]{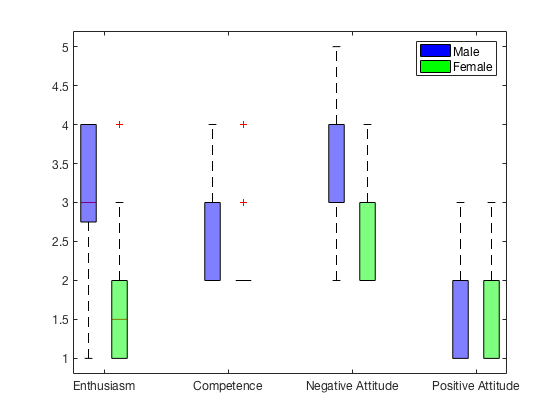}
\caption{TA-EG scores for males and females}\label{fig:bp_gender}
\end{figure}
Fig.~\ref{fig:bp_gender} shows the affinity ratings broken down by gender to explore how the gender is related to the technical affinity in our sample.
It is clear to see that in most cases the difference is small, also on the competence dimension, only the enthusiasm value is higher for men.
Again, for every group, low and high affinity, we performed separate $\chi^2$ goodness-of-fit tests against a uniform distribution. The $\chi^2$ test for the high affinity group yields results similar to our first test for all participants: in all categories the color choice is significantly different from a uniform distribution, for \textit{Active} $\chi^2(5, N=102) = 112.59$, for \textit{Help needed} $\chi^2(5, N=102) = 33.18$, for \textit{Error} $\chi^2(5, N=102) = 152$, $p\textless.01$ in all cases. Similar results were obtained for the data of the low affinity group, the color preference is again not equally distributed: for \textit{Active} $\chi^2(4, N=66) = 109.46$, for \textit{Help needed} $\chi^2(5, N=66) = 21.82$, for \textit{Error} $\chi^2(5, N=66)$ $ = 34$, $p\textless.01$ in all cases.
In order to compare the groups with each other, we performed a $\chi^2$ test of homogeneity for the different categories. In two out of three categories, we cannot report any significant difference between the groups since more than 20\% of the cells had an expected cell count smaller than 5. Thus, an assumption of the test could not be met. However, in the \textit{Help needed} category we can report a significant difference between the affinity groups ($\chi^2(5, N=168)=11.73$, $p=0.039$). The color red is chosen by 15.6\% in the high affinity group and by 31.8\% in the low affinity group. So people with a high affinity do not consider red as suitable as those with low affinity do. The color orange, on the contrary, is chosen most often for \textit{Help needed} in the high affinity group. In other words, the difference between color preferences for \textit{Help needed} and \textit{Error} is bigger for high affinity participants. The results are shown in Fig.~\ref{aff}.
\begin{figure}
  \begin{subfigure}{0.5\textwidth}
  \centering
  \includegraphics[width=0.8\textwidth]{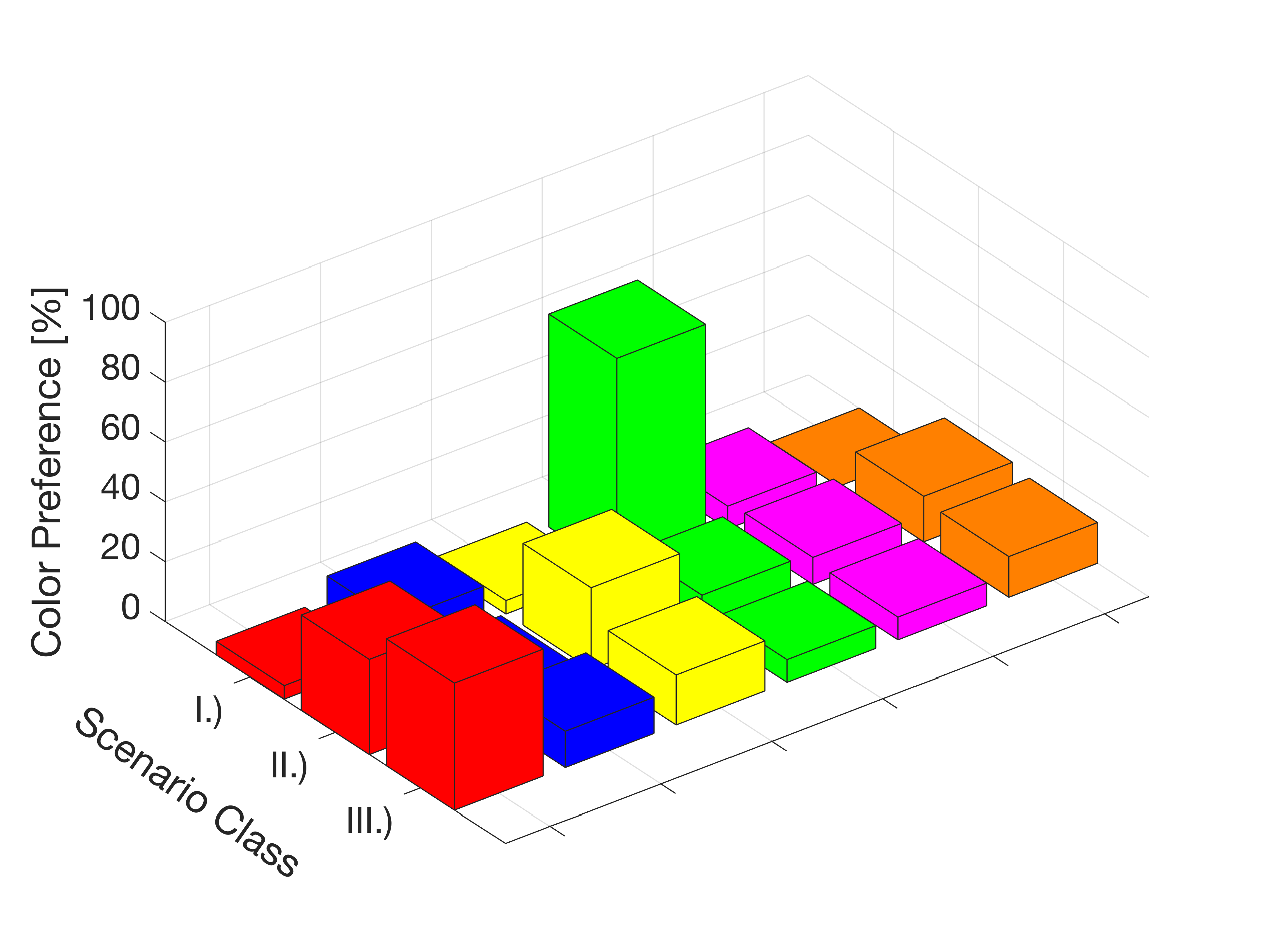}
  \subcaption{Color preferences in low affinity group}\label{lowaff}
  \end{subfigure}
\begin{subfigure}{0.5\textwidth}
\centering
\includegraphics[width=0.8\textwidth]{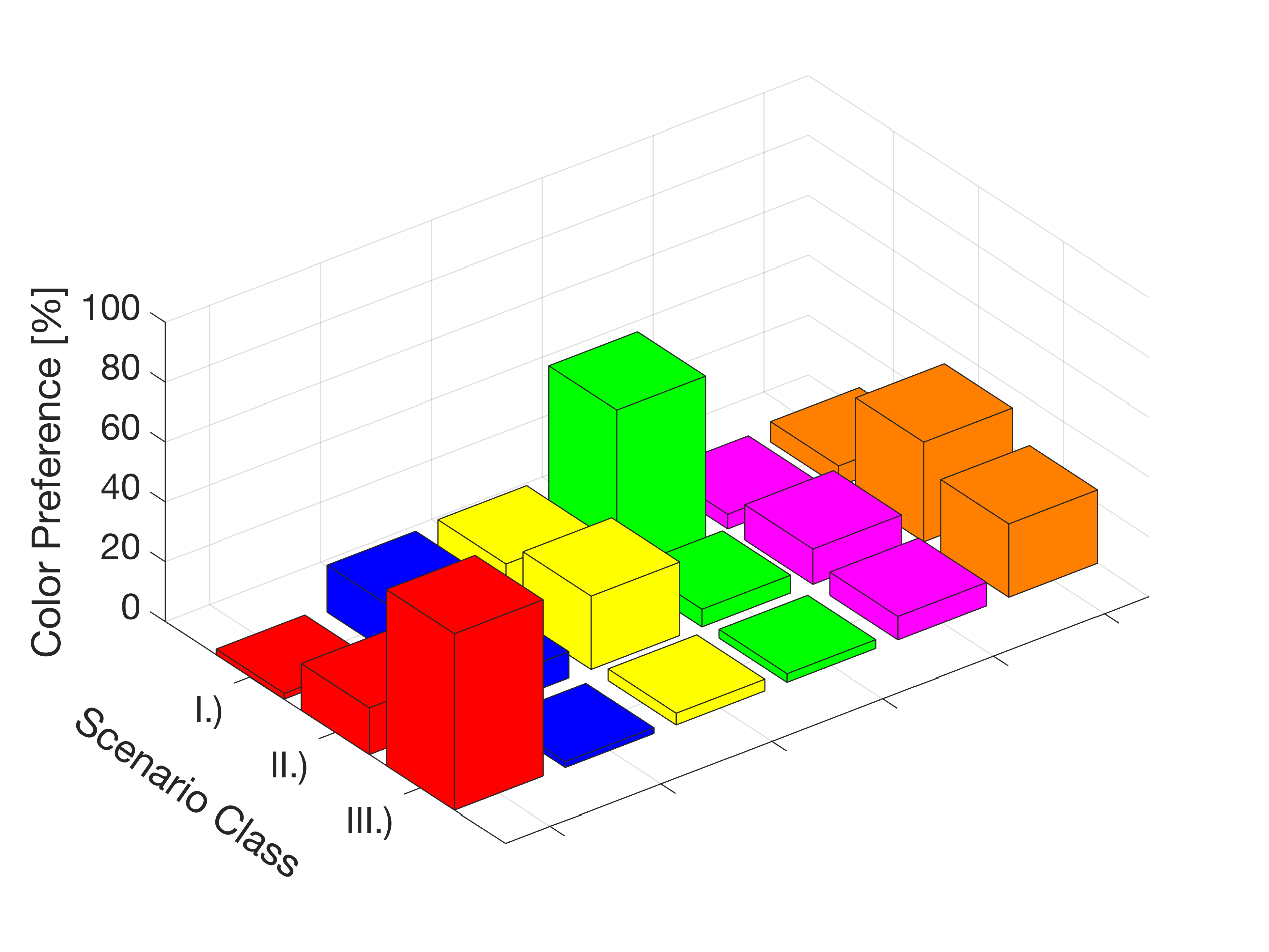}
\subcaption{Color preferences in high affinity group}\label{highaff}
\end{subfigure}
\vspace{-0.8em}\caption{Color preference considering the technical affinity}
\label{aff}
\end{figure}
\begin{figure}
  \begin{subfigure}{0.5\textwidth}
  \centering
  \includegraphics[width=0.8\textwidth]{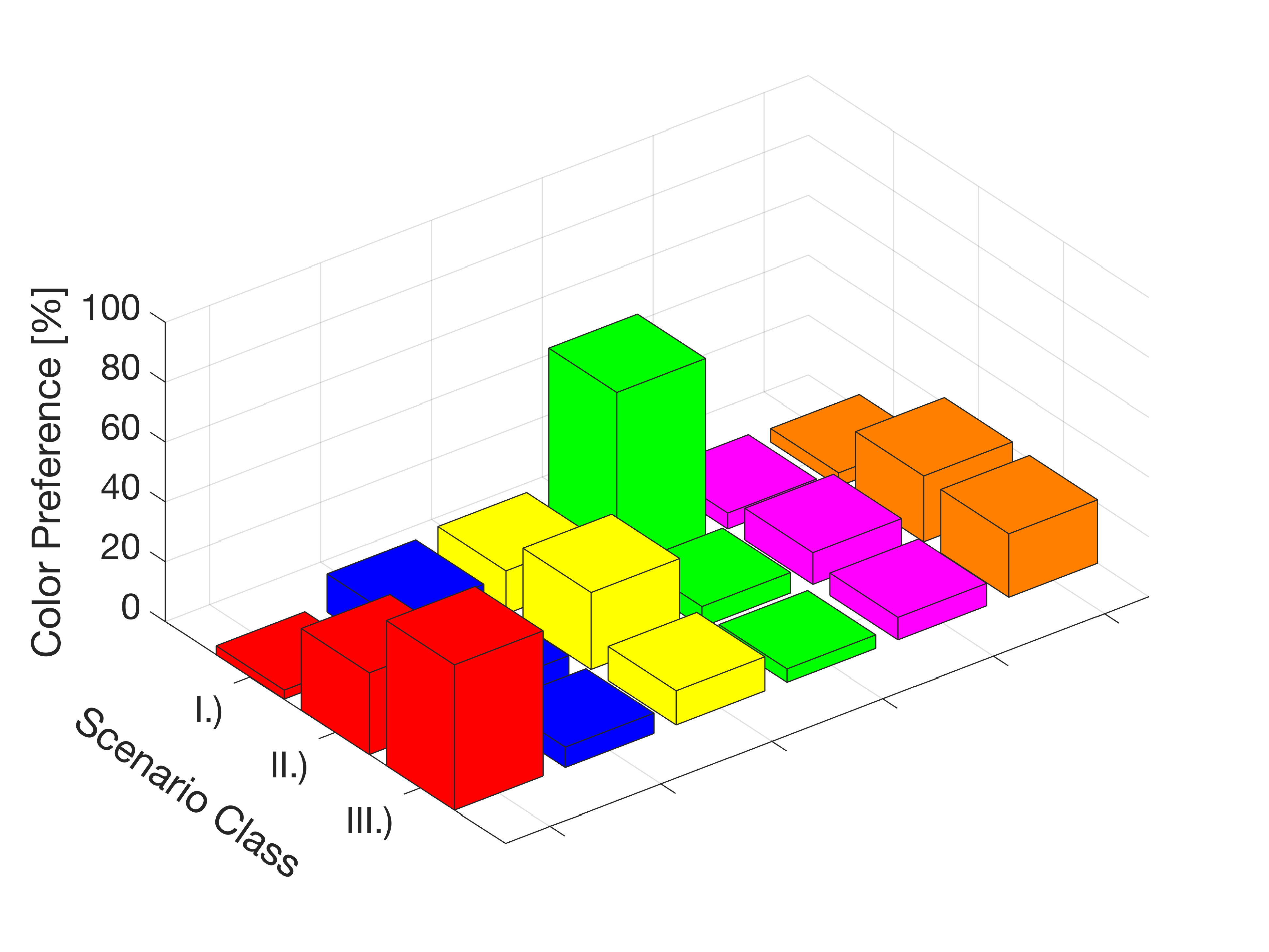}
  \caption{Robot non-experienced group}\label{robounexp}
  \end{subfigure}
\begin{subfigure}{0.5\textwidth}
\centering
\includegraphics[width=0.8\textwidth]{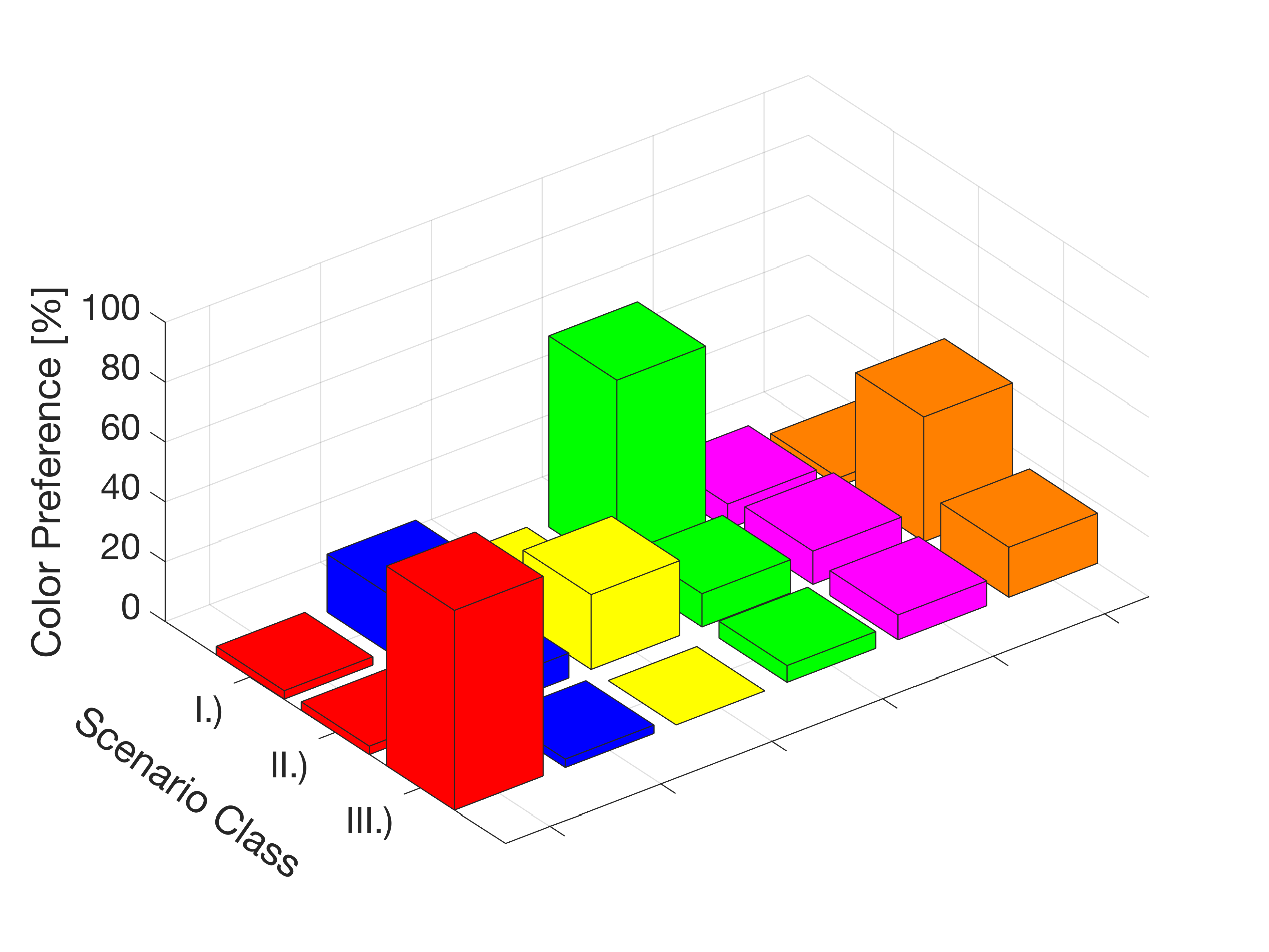}
\subcaption{Robot experienced group}\label{roboexp}
\end{subfigure}
\vspace{-0.8em}\caption{Color preference considering the robot experience}
\end{figure}

Moreover, we tested whether extended robot experience changes the results. Only 12 participants indicated they have contact to robots regularly. The votes of the robot experienced group as well as the robot non-experienced group are visualized in Fig. \ref{roboexp} respectively Fig.~\ref{robounexp}. Just like in all previous $\chi^2$ goodness-of-fit tests against a uniform distribution, the results for both groups, experts and non-experts, are significant, i.e. the choice is not random. In detail the test yields the following results for robot experts: for \textit{Active} $\chi^2(5, N=36) = 62.33$, for \textit{Help needed} $\chi^2(5, N=36) = 22$, for \textit{Error} $\chi^2(4, N=36) = 50.94$, $p\textless.01$ in all cases.
For the non-expert group we report for \textit{Active} $\chi^2(5, N=132) = 185.81$, for \textit{Help needed} $\chi^2(5, N=132) = 34.81$, for \textit{Error} $\chi^2(5, N=132) = 109.91$, $p\textless.01$ in all cases.
The dominant colors for \textit{Active} and \textit{Error} are the same as in the previous results, but we can observe a difference for the \textit{Help needed} category. As in the overall group, non-experts prefer red, orange and yellow. The experts also show high percentages for yellow and orange, but they do not choose red. The color red is chosen by only 2.8\% of the robot experts and by 27.3\% of the non-experts. The color orange, on the contrary, is chosen for the \textit{Help needed} category most often in the robot experts group. Both differences are similar to the results of the affinity analysis. Because in this case the requirements of the $\chi^2$ test of homogeneity are violated, we cannot report valid results.
\begin{figure}
  \begin{subfigure}{0.5\textwidth}
  \centering
  \includegraphics[width=0.8\textwidth]{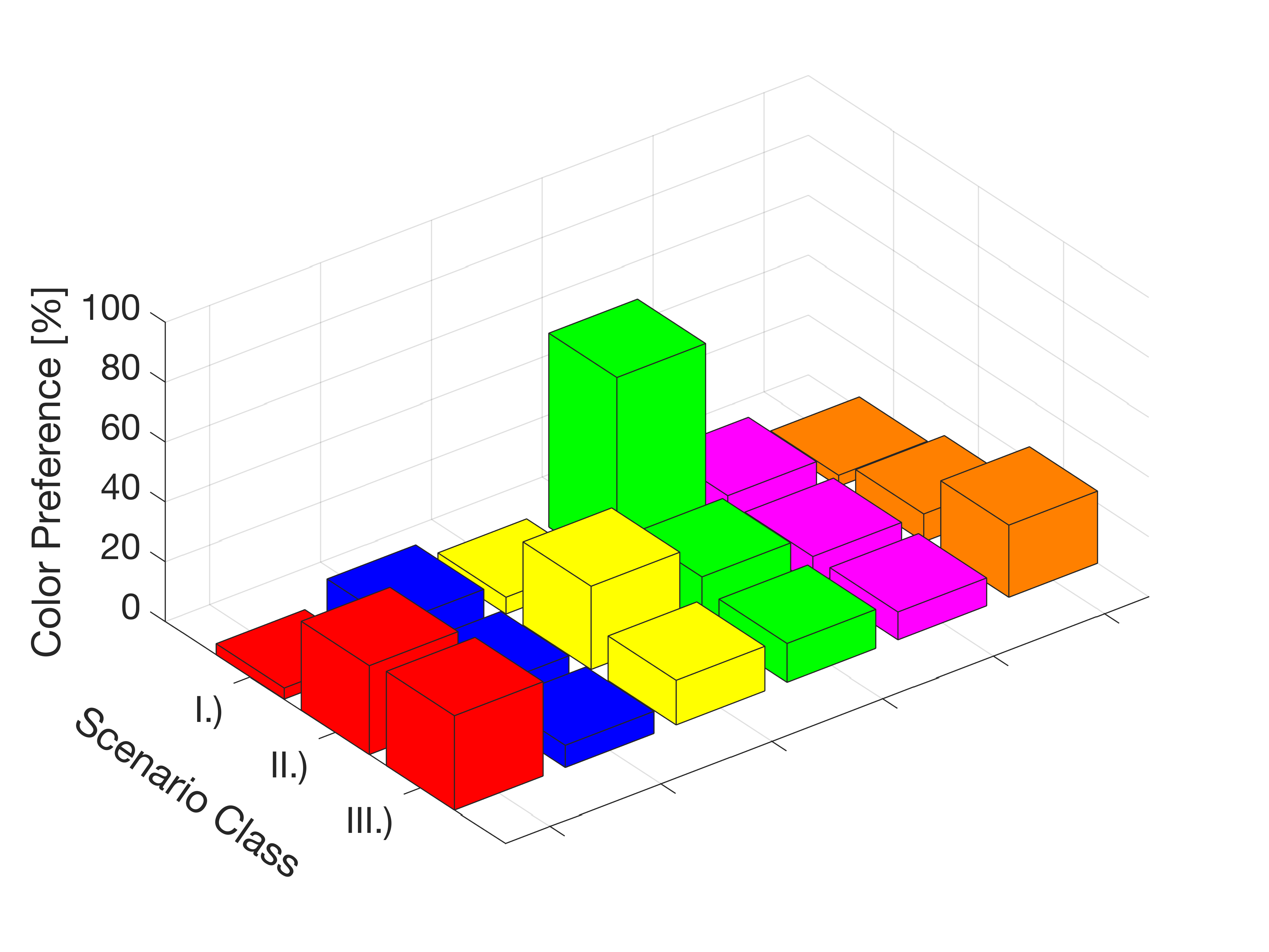}
  \subcaption{Color preferences in group of women}\label{fig:color_women}
  \end{subfigure}
\begin{subfigure}{0.5\textwidth}
\centering
\includegraphics[width=0.8\textwidth]{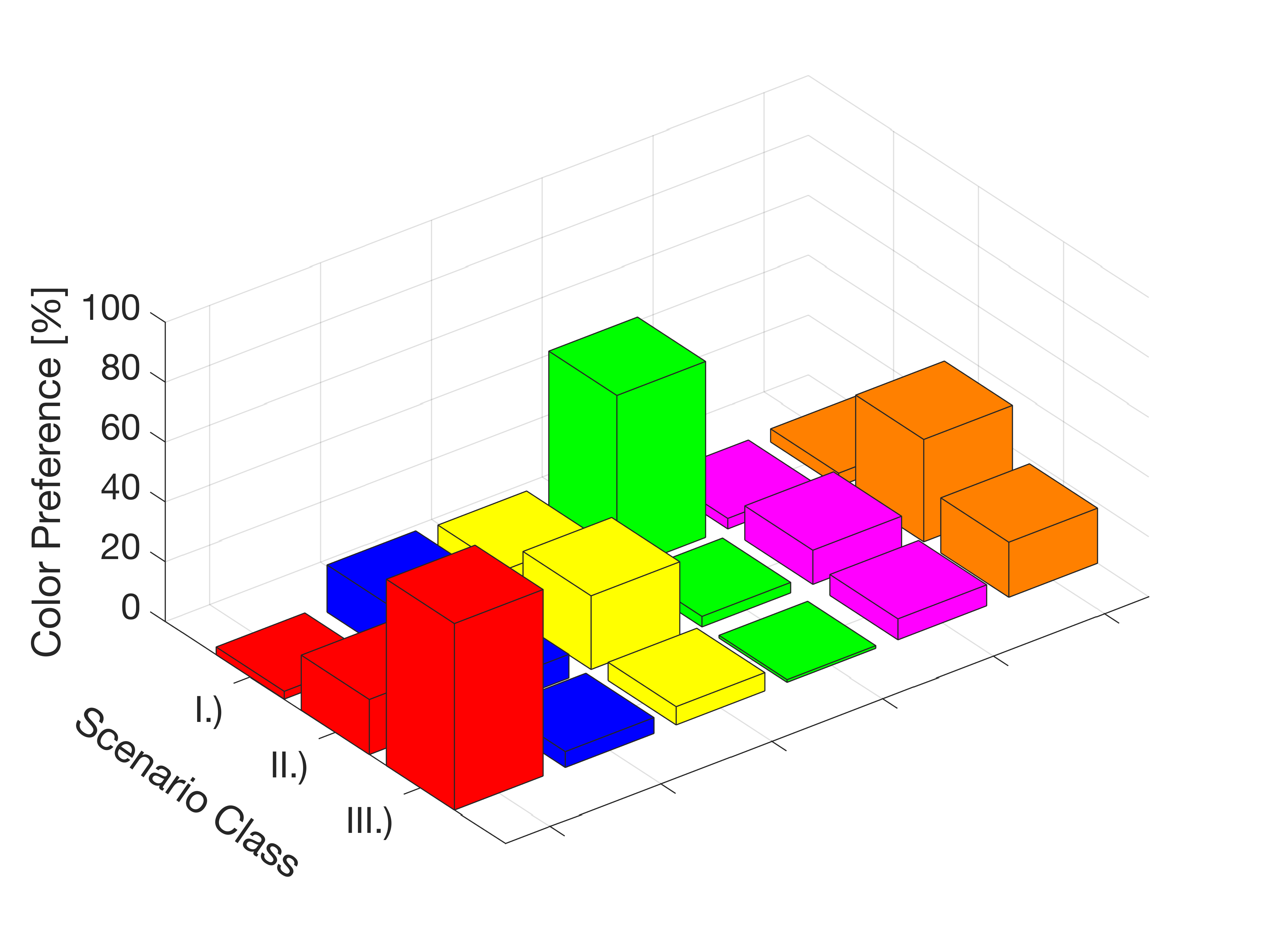}
\subcaption{Color preferences in group of men}\label{fig:color_men}
\end{subfigure}
\vspace{-0.8em}\caption{Color preference considering gender}
\end{figure}

Furthermore, we also consider the differences between men and women. The results are illustrated in Fig.~\ref{fig:color_women} and Fig.~\ref{fig:color_men}. While \textit{Active} is in both groups dominated by green, the distinction between \textit{Help needed} and \textit{Error} in the group of women seems ambiguous. 31.5\% of the women chose red for \textit{Error} but simultaneously 29.7\% of women chose red for \textit{Help needed}. Another predominant color for \textit{Error} in the group of women is orange which is chosen by 24.1\% of the female participants. Additionally, while 58.8\% of the group of males chose yellow and orange as an appropriate color for \textit{Help needed}, only 37.1\% of the women chose yellow or orange as an appropriate color.
The $\chi^2$ goodness-of-fit tests for each group indicate clearly that the color preference is not equally distributed: for \textit{Active} $\chi^2(5, N=114) = 157.16$ (male) and $\chi^2(5, N=54) = 92$ (female), $p\textless.01$ in both cases; for \textit{Help needed} $\chi^2(5, N=114) = 44.53$ (male) and $\chi^2(5, N=54) = 15.78$ (female), $p\textless.01$ in both cases; for \textit{Error} $\chi^2(5, N=114) = 182.42$, $p\textless.01$ (male) and $\chi^2(5, N=54) = 14$, $p = 0.16$ (female). Performing a $\chi^2$ test of homogeneity, we identified a significant difference between both groups concerning the \textit{Help needed} category ($\chi^2(5, N=168)=19.32$, $p=0.002$).

\section{Discussion}
The results show that colors are connected to semantic meanings, because the color preferences differ between different category.
We consider the consistency of the answers as a hint that the participants understood the videos correctly and could differentiate between the categories. Therefore, we conclude our first hypothesis \textbf{H1} to be supported. We do not know in each case whether the participant's interpretation of a video matches our intended category, which is a limitation of this study. Furthermore, the study is limited to a group of people who are socialized in Europe. This is important because the socialization can have an effect on the color semantics. Investigating this effect would be an interesting objective for further research.
In our survey, we offered six different color options. A possible expansion of the experiment is to not make a pre-selection of colors at all. To be able to conclude statements from the evaluation, this would require a big number of participants, which was not feasible in this study.
We hypothesized that the favorite color might influence the choices, but the results do not support this. If participants preferred their favorite color over others, this is only true for the \textit{Active} category, but the influence seems to be negligible. Therefore, we reject \textbf{H2}. The hypothesis \textbf{H3} assumes that the technical affinity of the participants influences their color selection. Since the evaluation for the \textit{Help needed} category shows a significant difference between the affinity groups, this hypothesis is supported by the results.
Equally, the difference between non-experts and experts in the \textit{Help needed} category is considerable, even if it is only visible for one category. The experience definitely influences the result, but it is not clear whether it influences the color semantics or only the understanding of the videos. Unfortunately, a statement about the statistical significance of the difference is not possible, due to the small number of robot experienced participants. It would be an interesting starting point for further research to examine this difference in detail.
For our sample, we are confident that \textbf{H4} is supported. Since the difference for the \textit{Help needed} category seems more evident comparing robot experienced with non-expert persons than that between low and high affinity, the robot experience might be the better control variable in this case. As mentioned earlier, only one dimension of the technical affinity is free from opinion and emotions, the competence, and the affinity groups do not differ much on this dimension. Probably, the attitude towards technical devices is less relevant than the extent of experience.
The gender differences in affinity occur on the emotion-related dimensions, but not on the competence dimension. This means the gender differences in color preferences cannot be explained by different levels of experience or affinity. Although we are not able to provide an explanation for the gender differences in color choice, we have to reject \textbf{H5}.
The fact that differences between color preferences only occur in one category might be a hint that the \textit{Help needed} category is either too ambiguous or not clearly connected to a color.
One possibility is that higher experience leads to stronger color-meaning associations. An alternative explanation is that the scenarios are easier to understand and the categories are easier to distinguish for experienced persons. This could explain why only the \textit{Help needed} category leads to different results: it is less obvious than a fatal error or cases where everything goes well.
Colors that are a good choice for experts interestingly also work for non-experts. Nevertheless, additional information should be included when designing feedback, especially for scenarios where a robot needs help, because inexperienced users could be confused otherwise.

\section{Conclusion}
The aim of this study was to examine the semantics of colors in a human-robot interaction scenario. In an online survey, we presented scenarios in the form of videos to participants who where asked to select a color. Throughout all our evaluations, it was clear to see that the color preferences for the scenarios are not evenly distributed amongst the colors. The results therefore support our assumption that colored light signals are suitable as a feedback mechanism for mobile robots. Even though we cannot offer a clear recommendation for every category, our results can be used as orientation for a robot's feedback design. If the robot is \textit{Active}, green light signals are most appropriate, if it experiences an \textit{Error}, we recommend red signals. If it needs \textit{Help} to accomplish a task, the signal should be in the color range between orange and yellow.
Additionally, the study revealed differences in the color selection between groups of people with high versus low technical affinity. The same applied - with limitations - to robot experienced users and robot non-experienced users. Finally, the study revealed gender-specific differences in the color selection that cannot be explained based on the present data.
The study has some limitations: we do not know for sure if people interpreted the videos correctly, even if the present data indicates it. In addition, the color choice had to be limited and the group of participants was limited to Europeans. Future studies with even more participants could use a color wheel widget to allow a free choice and clarify the connection between gender respectively robot experience and color. Our ideas could furthermore be extended to other information categories and a combination of light with additional modalities.
\newline
\bibliographystyle{IEEEtran}
\bibliography{lit}

\end{document}